# Beacon-Based Feedback Control for Parking an Active-Joint Center-Articulated Mobile Robot

*Mehdi Delrobaei, Kenneth A. McIsaac*

The University of Western Ontario
Department of Electrical and Computer Engineering
London, ON N6A5B9, Canada

## ABSTRACT

This paper presents an autonomous parking control strategy for an active-joint center-articulated mobile robot. We first derive a kinematic model of the robot, then propose a control law to stabilize the vehicle's configuration within a small neighborhood of the goal. The control law, designed using Lyapunov techniques, is based on the robot's polar coordinate equations. A beacon-based guidance system provides feedback on the target's position and orientation. Simulations demonstrate the robot's ability to park successfully from arbitrary initial poses.

## 1. MOTIVATION

*Modular robotic systems*, inherently robust and flexible, continue to challenge researchers [1]. Robots that can change shape are built for chain, lattice, or mobile reconfiguration. Mobile self-reconfiguring robots change shape by having modules detach themselves from the main body and move independently. They then link up at new locations to form new configurations. This type of reconfiguration is less explored than the other two because the difficulty of reconfiguration tends to outweigh the gain in functionality [2].

In the majority of current implementations, modular robots, even those that can self-reconfigure, are initially manually assembled [1]. Thus, there is an opportunity to develop an autonomous modular system which disassembles into a set of independent and autonomous mobile units to spread out in an area, and later reassembles into the single robot.

The overall objective of our research work is to develop an autonomous multi-rover robot, in which a team of autonomous articulated-steering mobile robots can assemble into a single serial-chain modular robot. In our previous work [3], we presented the main idea as well as the design and implementation of a reconnectable joint which is needed to connect and disconnect the modules. In this paper, we propose a parking (docking) control system, needed to steer independent modules such that they are prepared to join to one another.

## 2. INTRODUCTION

In *articulated steering*, the heading of the robot changes by folding the hinged chassis units. In contrast with other types of mobile robots, literature on feedback control of center-articulated mobile robots is not extensive. The reason might be that application of articulated-steering mechanism is limited to heavy duty vehicles for mining applications [4].

Apostolopoulos [5] presented a practical analytical framework for synthesis and optimization of wheeled robotic locomotion configurations, including articulated-steering type vehicles. Choi and Sreenivasan [6] investigated the kinematics of a multi-module vehicle using a numerical approach. The number of actuators in this design can vary from nine (four for wheels, two for variable-length axles, and three for the articulation) in a fully active system to a minimum of three. Azad et al. [7] investigated the dynamic behavior of an articulated steering vehicle.

Load-haul-dump (LHD) vehicles which are typically used in underground mines are articulated-steering vehicles, and their steering kinematics resembles center-articulated mobile robots kinematics. Corke and Ridley [8] developed a general kinematics model of the vehicle that shows how heading angle evolves with time as a function of steering angle and velocity. A path-tracking criterion for LHD trucks is proposed in [9]. Marshall [10] *et al.* have also investigated localization and steering of an LHD vehicle in a mining network.

In another work, Ridley and Corke [11] derived a linear, state-space, mathematical model of center-articulated vehicles, purely from geometric consideration of the vehicle and its desired path. Then, autonomous regulation of the vehicle is shown to be theoretically feasible using state variable feedback of displacement, heading, and curvature error.

The parking problem basically involves formulating a trajectory for a given object or system of objects, such that predefined geometric and/or non-geometric constraints are satisfied. Choset *et al.* [12] studied motion planning for non-

This work was supported by the Natural Sciences and Engineering Research Council (NSERC) of Canada under Grant RGPIN 249883-06.

holonomic and under-actuated systems. They have investigated the controllability problem, and have focused on developing motion planning algorithms. Laumond [13] studied motion planning for non-holonomic mobile robots focusing on car-like robots and probabilistic path planning. Ahrikenchenikh [14] investigated optimized motion planning algorithms for such robots.

Ishimoto *et al.* [15] considered the problem of trajectory following of a wheeled loader drawing a V-shape path when it picks up gravel and loads it to a dump truck. A symmetrical clothoid, which is described as a function of a distance parameter, is proposed to specify the V-shape trajectory. They illustrated a method to generate the clothoid and an implementation on a miniature wheel loader.

In this paper, a closed-loop parking control system for articulated-steering mobile robots is proposed, based on Lyapunov stability theory. To achieve feedback control, the goals information is needed. In this work, a positioning system is introduced in which bearing from three active beacons provide enough information to uniquely determine the position and orientation of the target.

The paper is organized as follows. In Section 3, the kinematic model of a center-articulated mobile robot is presented. In Section 4, the design of the controller is illustrated. Section 5 describes the positioning system. Section 6 reports the simulation results, and Section 7 contains some conclusion and future work.

## 3. KINEMATIC MODEL

A center-articulated mobile robot consists of two bodies joined by an active joint. The vehicle is steered by changing the body angle.

Consider an active-joint center-articulated mobile robot positioned at a non-zero distance with respect to a target frame (Fig. 1). The robot's motion is governed by the combined action of the linear velocity $v$ and the angular velocity $w$.

The kinematic equations of the robot which involve the robot's Cartesian position $(x, y)$ and the heading angle of the front body $\psi$ can be written as:

$$\dot{x} = v\cos\psi \quad (1)$$

$$\dot{y} = v\sin\psi \quad (2)$$

$$\dot{\psi} = \frac{\sin\phi}{l_2 + l_1\cos\phi}v + \frac{l_2}{l_2 + l_1\cos\phi}\omega \quad (3)$$

$$\dot{\phi} = \omega \quad (4)$$

where $l_1$ and $l_2$ are the lengths of the front and the rear parts of the robot, and $\phi$ is the body angle.

It should be noted that to write these equations it is assumed that the velocity vector of the front body is orthogonal to the front wheels' axle (no side-slips).

The kinematic equations can also be written in polar coordinates as:

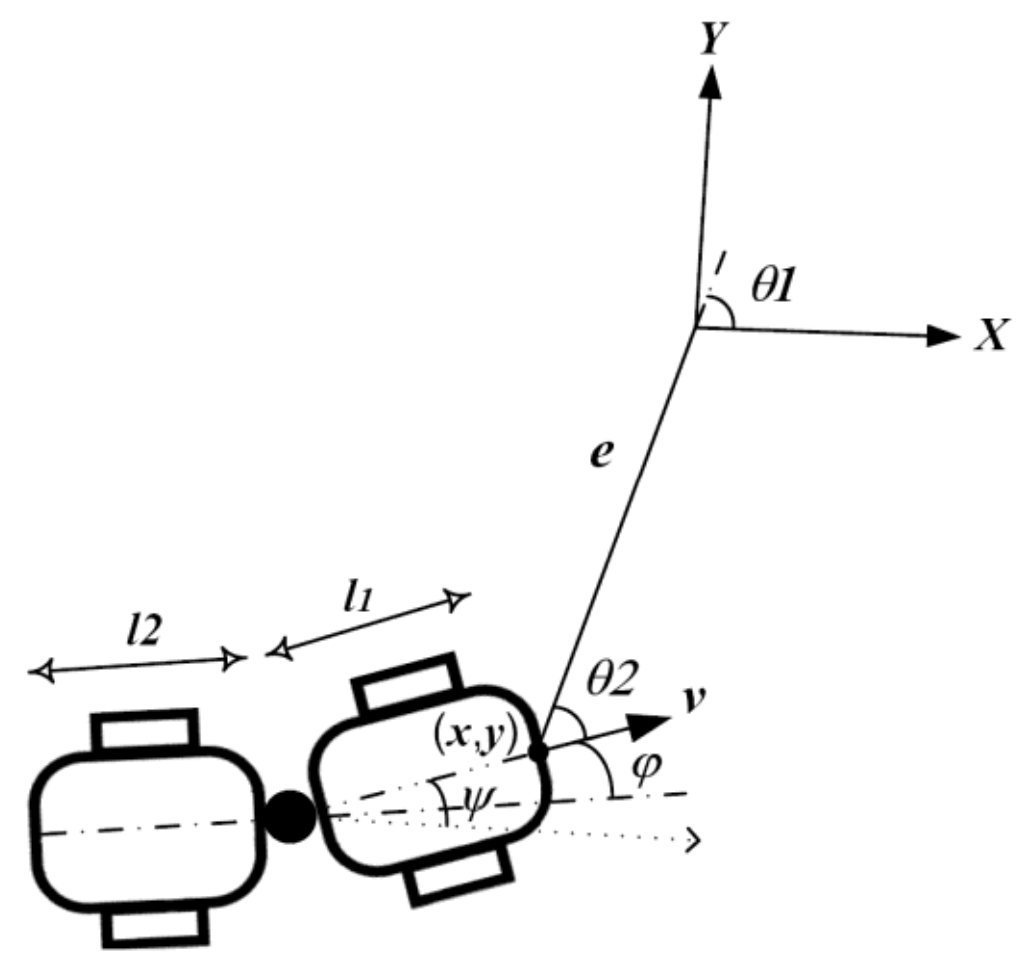

**Fig. 1**. Diagram of a center-articulated mobile robot with respect to the target frame. It is shown that using polar coordinates is suitable to design an appropriate control law for parking maneuver of center-articulated mobile robots.

$$\dot{e} = -v\cos\theta_2 \quad (5)$$

$$\dot{\theta}_1 = \frac{v\sin\theta_2}{e} \quad (6)$$

$$\dot{\theta}_2 = \left(\frac{\sin\theta_2}{e} - \frac{\sin\phi}{l_2 + l_1\cos\phi}\right)v - \frac{l_2}{l_2 + l_1\cos\phi}\omega \quad (7)$$

$$\dot{\phi} = \omega \quad (8)$$

where $e$ is the error distance, $\theta_1$ is the error vector orientation with respect to the target frame, and $\theta_2$ is the angle between the distance vector $e$ and the linear speed vector.

It is interesting to note that using polar coordinates allows for a set of state variables which closely resemble the same ones regularly used within our car-driving experience [16]. In the next section, it will be shown that equations (5)-(8) are suitable to design an appropriate control law for parking maneuvers.

## 4. PARKING CONTROL

In this section, first the steering controls are determined and then the Lyapunov analysis is carried out.

### 4.1. Controller Design

The Lyapunov stability theory is a common tool to design control systems. Here we consider a simple quadratic equation as a candidate Lyapunov function.

Let the robot be initially positioned at a non-zero distance from the target frame. The objective of the parking control system is to move the robot so that it is accurately aligned with the target frame.

In other words, it is intended to find a stable control law $[v(e, \theta_1, \theta_2, \varphi),\ w(e, \theta_1, \theta_2, \varphi)]$ which drives the robot from any initial position $(e(0), \theta_1(0), \theta_2(0))$ to a small neighborhood of the target, (0,0,0).

Consider the positive definite form

$$V = \frac{1}{2}\lambda_1 e^2 + \frac{1}{2}\lambda_2\theta_1^2 + \frac{1}{2}\lambda_3\theta_2^2 + \frac{1}{2}\lambda_4\phi^2 \qquad (9)$$
$$\lambda_1, \lambda_2, \lambda_3, \lambda_4 > 0$$

The time derivative of $V$ can be expressed as

$$\dot{V} = \lambda_1 e\dot{e} + \lambda_2\theta_1\dot{\theta}_1 + \lambda_3\theta_2\dot{\theta}_2 + \lambda_4\phi\dot{\phi} \qquad (10)$$

Substituting (5)-(8) in (10) gives

$$\dot{V} = [(\lambda_2\theta_1 + \lambda_3\theta_2)\frac{\sin\theta_2}{e} - \lambda_1 e\cos\theta_2 - \frac{\lambda_3\theta_2\sin\phi}{l_2 + l_1\cos\phi}]v + (\lambda_4\phi - \frac{l_2\lambda_3\theta_2}{l_2 + l_1\cos\phi})\omega \qquad (11)$$

It can be seen that letting

$$v = -[(\lambda_2\theta_1 + \lambda_3\theta_2)\frac{\sin\theta_2}{e} - \lambda_1 e\cos\theta_2 - \frac{\lambda_3\theta_2\sin\phi}{l_2 + l_1\cos\phi}] \qquad (12)$$

and

$$\omega = -(\lambda_4\phi - \frac{l_2\lambda_3\theta_2}{l_2 + l_1\cos\phi}) \qquad (13)$$

makes $\dot{V} \leq 0$ which implies stability of the system states.

Convergence (asymptotic stability) depends on the choice of $\lambda$s, as discussed in the next section.

### 4.2. Stability Analysis

The proposed candidate Lyapunov function $V$ is lower bounded. Furthermore, $\dot{V}$ it negative semi-definite and uniformly continuous in time ($\ddot{V}$ is finite). Therefore, according to Barbalat's lemma [17], $\dot{V} \to 0$ as $t \to \infty$.

The time derivative of $V$ can be expressed as

$$\dot{V} = \dot{V}_1 + \dot{V}_2 = -[(\lambda_2\theta_1 + \lambda_3\theta_2)\frac{\sin\theta_2}{e} - \lambda_1 e\cos\theta_2 - \frac{\lambda_3\theta_2\sin\phi}{l_2 + l_1\cos\phi}]^2 - (\lambda_4\phi - \frac{l_2\lambda_3\theta_2}{l_2 + l_1\cos\phi})^2 \qquad (14)$$

If $\lambda_4$ is selected to be very small, $\dot{V}_2$ takes on the form

$$\dot{V}_2 \approx -(\frac{l_2\lambda_3\theta_2}{l_2 + l_1\cos\phi})^2 \qquad (15)$$

So, $\dot{V}_2 \to 0$ implies that $\theta_2 \to 0$.

As $\theta_2 \to 0$, $\dot{V}_1$ also takes on a simpler form of

$$\dot{V}_1 \approx -(\lambda_2\theta_1\frac{\theta_2}{e} - \lambda_1 e)^2 \qquad (16)$$

Consequently, $\dot{V}_1 \to 0$ gives

$$\lambda_2\theta_1\theta_2 \approx \lambda_1 e^2 \qquad (17)$$

As $\theta_2 \to 0$, we get $e \to 0$.

Finally, in the limit where both $e$ and $\theta_2$ go to zero, $\theta_2/e$ is bounded and (12) gives

$$\lim_{e,\theta_2\to 0} v = -\frac{\lambda_2\theta_1\theta_2}{e} \qquad (18)$$

Therefore, from (6)

$$\dot{\theta}_1 = -\lambda_2(\frac{\theta_2}{e})^2\theta_1 \qquad (19)$$

As $\lambda_2 > 0$ and $(\frac{\theta_2}{e})^2$ is positive, from (19) it is found that $\theta_1$ is stable and eventually approaches zero, though it may do so slowly.

Therefore, $\dot{V} \to 0$ results in $(e, \theta_1, \theta_2) \to (0, 0, 0)$. It is noted that this analysis does not imply that $\phi \to 0$. This point should not cause any problem for the robot's parking maneuver since $\phi$ is only the body angle. It can be easily adjusted after the robot is docked.

In practice, there is a trade-off in selecting parameter $\lambda_4$. Setting $\lambda_4$=0 stabilizes $(e, \theta_1, \theta_2)$ while rendering $\phi$ uncontrollable. In such cases $\phi$ can take on physically unrealizable values, for example, causing the robot to fold in on itself. By contrast, choosing $\lambda_4$ large can result in very slow approaches to the origin.

It should also be mentioned that the proposed model for the center-articulated mobile robots has a singularity at $e = 0$, since according to (6) and (7), $\dot{\theta}_1$ and $\dot{\theta}_2$ are not defined at $e = 0$. This point that distance error $e$ is monotonically non increasing in time ($e \to 0$ as $t \to \infty$) indicates that condition $e = 0$ cannot occur in finite time.

One may also observe another singularity. If $l_2 + l_1 cos\phi = 0$ then $\theta_2$ is not defined. If the robot is designed such that $l_2 > l_1$, this singularity never happens. If $l_2 = l_1$, $cos\phi = \pm\pi$ results in this singularity, but, this case cannot occur since it means that the robot is fully folded back on itself.

Finally, we note that there is a special case where the controller is not able to stabilize the configuration of the robot. This special case occurs when both $\phi$ and $\theta_2$ are initially zero. As can be observed from (12) and (13), in this situation $\omega = 0$ and $v = \lambda_1 e$. In fact, there is no control on $\theta_1$. The controller should recognize this special case and take a proper action. For instance, the controller can change the initial body angle to a non-zero value.

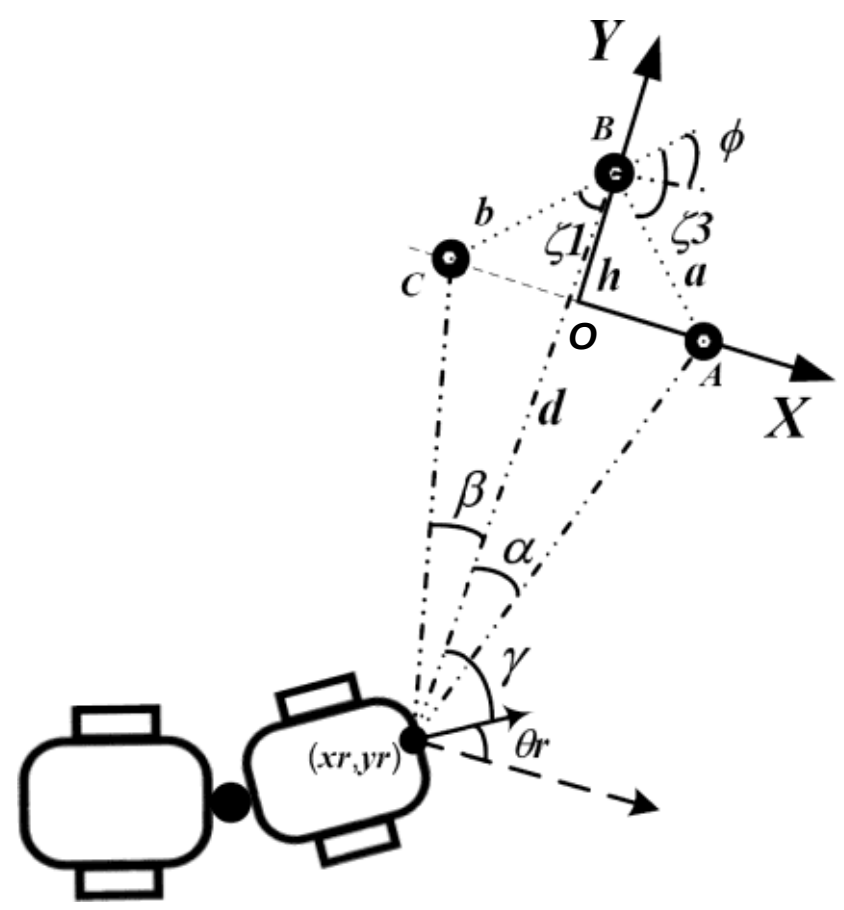

**Fig. 2**. Given three active beacons on the target frame and the robot with the ability to only detect the angles between them, the robot's position and orientation can be determined.

## 5. POSITIONING

In order to achieve closed-loop control based on control law described in (12), (13), feedback from position and orientation is needed. Considering practical considerations, measuring the distance and the robot's heading angle is not trivial. In this section, we propose a method to provide the feedback only based on bearing from active beacons.

Referring to Fig. 2, the beacons are labeled *A*, *B*, and *C*. The beacons locations are known (with respect to the target frame), and angles $\alpha$, $\beta$, and $\gamma$ are measured. The objective of the positioning system is to determine $e, \theta_1, \theta_2$ (see Fig. 1).

According to Fig 2, the robot's position $(x_r, y_r)$ and orientation $\theta_r$ are

$$x_r = -d\cos(\varphi + \zeta_1) \tag{20}$$

$$y_r = h - d\sin(\varphi + \zeta_1) \tag{21}$$

$$\theta_r = \varphi + \zeta_1 - \gamma \tag{22}$$

where $\varphi$, angle between line $\overline{CB}$ and X-axis, and *h*, Y-coordinate of beacon *B*, are known and depend on the position of the beacons. Therefore, it is necessary to calculate *d*, the distance from robot's heading and beacon B, and $\zeta_1$, the angle form vector *d* and line *CB*, to determine $(x_r, y_r, \theta_r)$.

From the law of sines we can write:

$$\triangle OAB: \ \frac{a}{\sin\alpha} = \frac{d}{\sin(\zeta_1 + \zeta_3 - \alpha)} \tag{23}$$

$$\triangle OCB: \ \frac{b}{\sin\beta} = \frac{d}{\sin(\beta + \zeta_1)} \tag{24}$$

where *a* and *b* are the distances between the beacons, and $\zeta_3$ is the beacons' exterior angle (see Fig. 2).

Dividing (23) by (24) and expanding the sine terms we obtain:

$$\zeta_1 = \tan^{-1}\left(\frac{b\sin\alpha\sin\beta - a\sin\beta\sin(\zeta_3 - \alpha)}{a\sin\beta\cos(\zeta_3 - \alpha) - b\sin\alpha\cos\beta}\right) \tag{25}$$

Having $\zeta_1$, distance *d* can be easily derived from (24),

$$d = \frac{b\sin(\beta + \zeta_1)}{\sin\beta} \tag{26}$$

Now that $(x_r, y_r, \theta_r)$ are determined (equations (20)-(22)), the configuration can be easily derived as

$$\begin{aligned} e &= \sqrt{x_r^2 + y_r^2} \\ \theta_1 &= \tan^{-1}\left(\frac{y_r}{x_r}\right) \\ \theta_2 &= \theta_1 - \theta_r \end{aligned} \tag{27}$$

It is noted that $\phi$ is the robot's body angle and can be directly measured by on-board sensors.

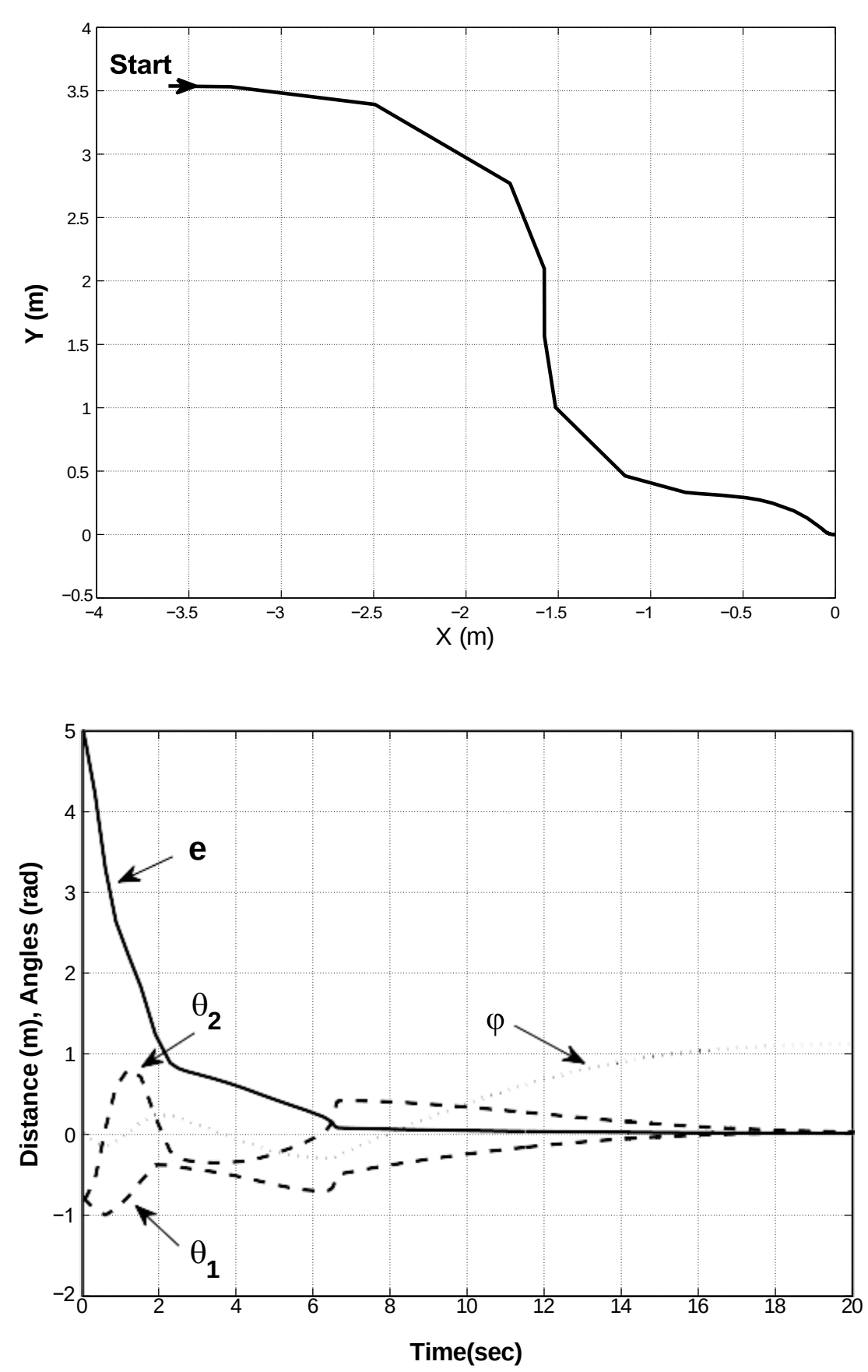

**Fig. 3**. The robot is initially positioned at (5,-$\pi$/4,-$\pi$/4,0); robot's trajectory (up), and state variables (down).

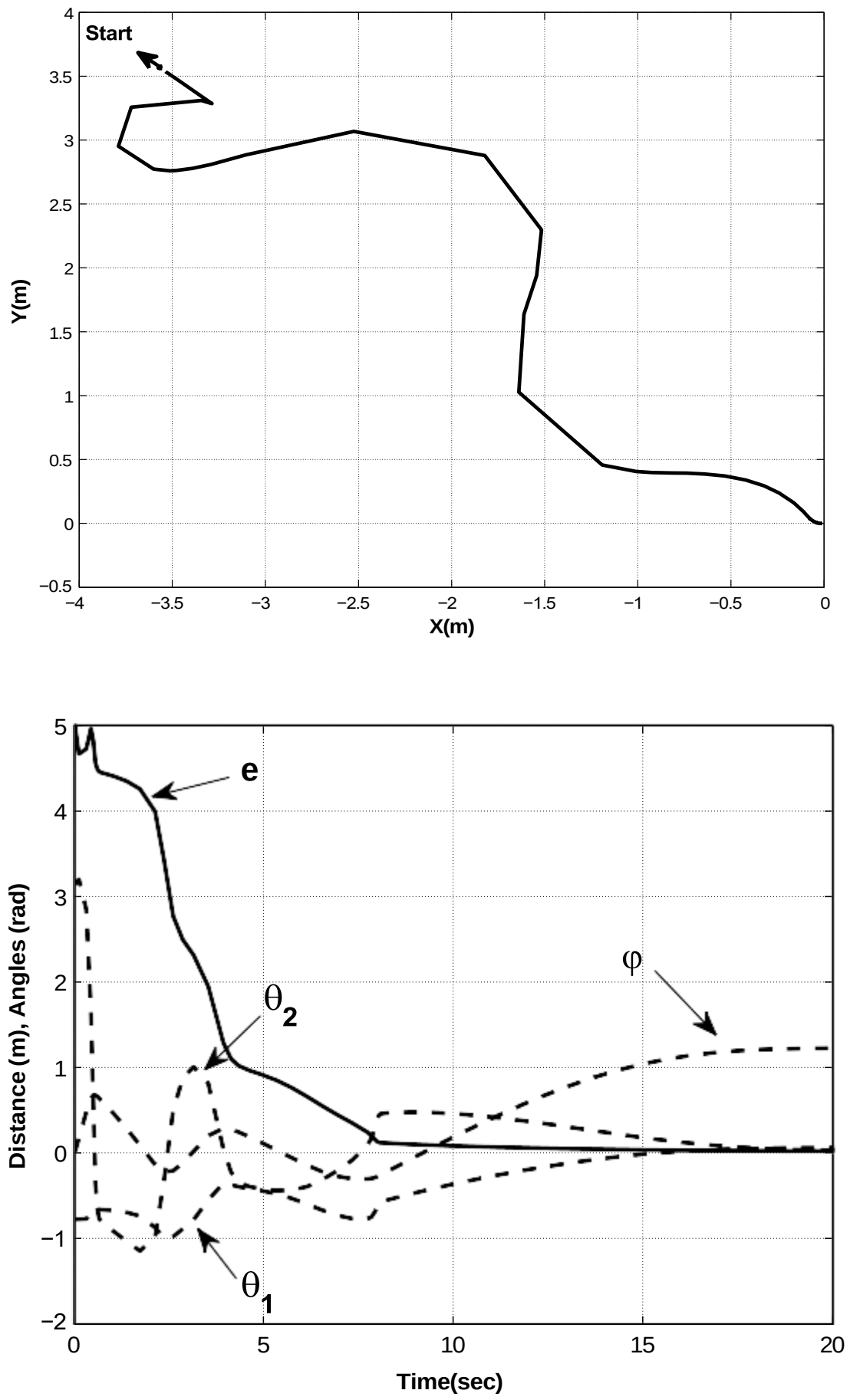

**Fig. 4**. The robot is initially positioned at (5,-$\pi$/4,$\pi$,0); robot's trajectory (up), and state variables (down).

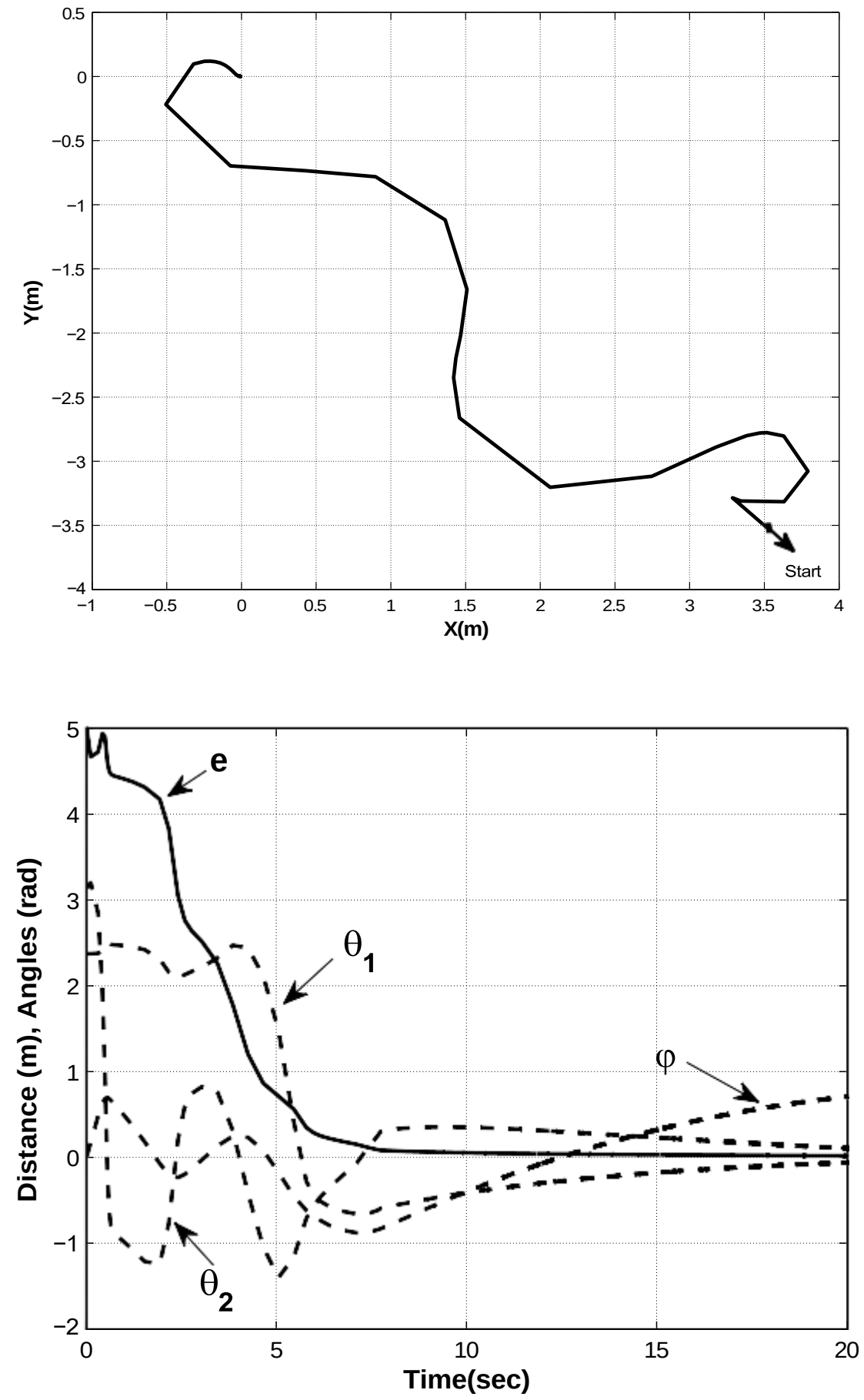

**Fig. 5**. The robot is initially positioned at (5,3$\pi$/4,$\pi$,0); robot's trajectory (up), and state variables (down).

## 6. SIMULATION RESULTS

Next, to evaluate the performance of the proposed controller, we present some simulation results. For our simulation, we suppose that the center-articulated robot is resting at different positions and orientations on X-Y plane (denoted in order by ($e$, $\theta_1$, $\theta_2$, $\varphi$)), and the target is the origin. We assume that the lengths of the robot are $l_1 = l_2 = 0.1$ m. The controller coefficients are set to $\lambda_1 = \lambda_2 = \lambda_3 = 1$ and $\lambda_4 = 0.01$.

In Fig. 3, the robot starts from (5,-$\pi$/4,-$\pi$/4,0). As it can be seen, the robot directly approaches the goal. In Fig. 4, the robot's initial position is (5,-$\pi$/4,$\pi$,0), so it is located at the same location as Fig. 3 but the heading angle is in the opposite direction. It is seen that the robot first moves backward and then moves towards the goal with positive velocity. The figures also show how the state variables converge.

In Fig. 5 and 6, the robot is initially resting at (5,3$\pi$/4,$\pi$,0) and (5,$\pi$,$\pi$,0), respectively. As can be seen, the robot approaches the target with an acceptable accuracy.

## 7. CONCLUSION AND FUTURE WORK

The autonomous parking problem of active-joint center-articulated mobile robots has been addressed in this paper. The results reveal that choosing a suitable state model allows to use a simple Lyapunov function to achieve parking control. Some simulation examples were illustrated to show the behavior of the robot during a parking maneuver. A positioning algorithm is used, based on bearing from active beacons, to provide feedback parameters.

Our future work includes integrating the proposed control system with the designed connection mechanism to implement an autonomous multi-rover robot. This modular robot changes shape by having its modules disassemble and move independently. They may autonomously reassemble into a new configuration in a later time.

The positioning system can be implemented using an omnidirectional camera and a set of color LEDs.

The overall objective is to introduce a new robotic concept which allows frail and inflexible mobile robots to convert to a single flexible modular robot with new capabilities, or a long and slow structure to split up into separate fast modules.

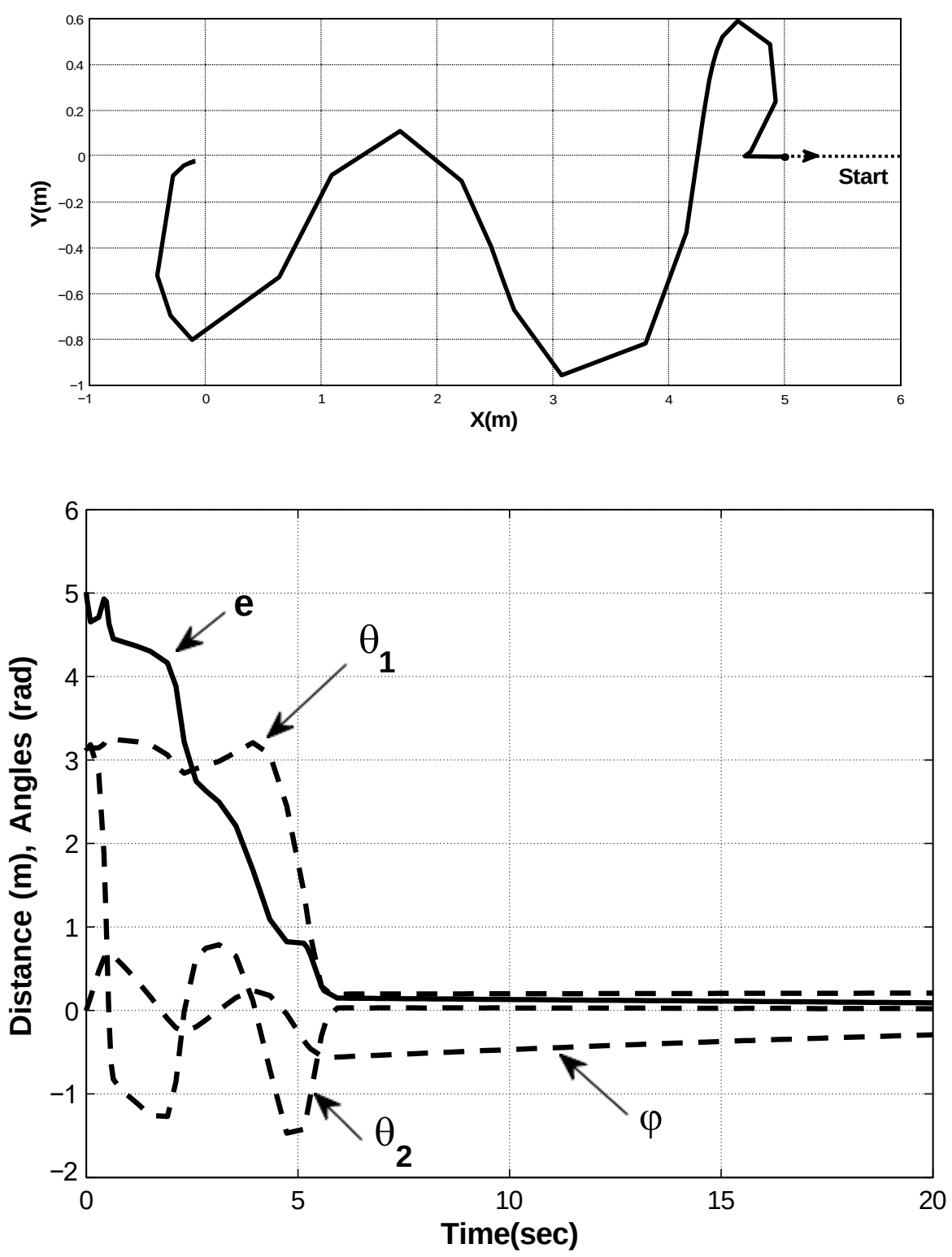

**Fig. 6**. The robot is initially positioned at (5,$\pi$,$\pi$,0); robot's trajectory (up), and state variables (down).